\documentclass{article}
\usepackage[utf8]{inputenc}
\usepackage{main}
\usepackage{microtype}
\usepackage{graphicx}
\usepackage{times}
\usepackage{amsmath}
\usepackage{amssymb}
\usepackage{enumitem}
\usepackage{booktabs}
\usepackage{natbib}
\definecolor{mydarkblue}{rgb}{0,0.08,0.45}
\usepackage[colorlinks=true,linkcolor=mydarkblue,citecolor=mydarkblue,filecolor=mydarkblue,urlcolor=mydarkblue]{hyperref}
\usepackage{fancyhdr}

\usepackage{xspace}
\usepackage{gradient-text}
\usepackage{float}
\usepackage{multirow}

\usepackage{url}
\usepackage{xcolor}
\usepackage[most]{tcolorbox}
\usepackage{fvextra}
\usepackage{subcaption}
\usepackage{tikz}
\usepackage{pgfplots}
\usepgfplotslibrary{groupplots}
\usetikzlibrary{matrix}
\pgfplotsset{compat=1.18}

\setlist[itemize,1]{leftmargin=10pt}

\usepackage{tikz}
\usepackage{soul}
\usepackage{colortbl}
\usepackage{subcaption}
\usepackage{bbm}
\usepackage{adjustbox}
\usepackage{makecell}
\usepackage[framemethod=TikZ]{mdframed}

\definecolor{mypositive}{RGB}{0, 128, 0}
\definecolor{mynegative}{RGB}{220, 20, 60}
\definecolor{mypositive}{RGB}{24, 103, 173}
\definecolor{mynegative}{RGB}{255, 128, 0}

\definecolor{customblue}{HTML}{1F77B4}
\definecolor{customorange}{HTML}{FF7F0E}
\definecolor{customgreen}{HTML}{2CA02C}
\definecolor{no}{HTML}{BA8E23}
\definecolor{darkgreen}{RGB}{0,100,0}
\definecolor{darkred}{RGB}{139,0,0}

\newcommand{\deltaScore}[1]{%
\begingroup
\ifdim #1pt > 0pt
\textcolor{darkred}{\scriptsize$^{#1}$}%
\else
\textcolor{darkgreen}{\scriptsize$^{#1}$}%
\fi
\endgroup
}

\newtcolorbox{boxblue}{enhanced,colback=blue!5!white,colframe=blue!75!black,breakable=true}

\usepackage{listings}
\usepackage{wrapfig}

\lstdefinelanguage{json}{
    basicstyle=\scriptsize\ttfamily,
    numbers=none,
    numberstyle=\tiny,
    stepnumber=1,
    numbersep=5pt,
    showstringspaces=false,
    breaklines=true,
    frame=none,
    backgroundcolor=\color{white},
    literate=
     *{:}{{{\color{black}{:}}}}{1}
      {,}{{{\color{black}{,}}}}{1}
      {\{}{{{\color{black}{\{}}}}{1}
      {\}}{{{\color{black}{\}}}}}{1}
      {[}{{{\color{black}{[}}}}{1}
      {]}{{{\color{black}{]}}}}{1},
}

\definecolor{YaleBlue}{RGB}{0, 53, 107}
\definecolor{UNCBlue}{RGB}{75, 156, 211}
\definecolor{ZJBlue}{RGB}{1, 126, 199}

\newcommand{\UNC}{\hspace{.1em}^{\textcolor{UNCBlue}{\boldsymbol{C}}}}
\newcommand{\Yale}{\hspace{.1em}^{\textcolor{YaleBlue}{\boldsymbol{Y}}}}

\newcommand{\data}{\textsc{\gradientRGB{STEPWISE}{64,64,64}{179,26,26}}}

\usepackage[flushmargin,hang]{footmisc}
\definecolor{darkblue}{rgb}{0, 0, 0.5}
\definecolor{stuckred}{RGB}{220, 50, 47}
\definecolor{milestonegreen}{RGB}{42, 161, 152}
\definecolor{codebg}{RGB}{253, 246, 227}
\definecolor{rowbg}{RGB}{238, 232, 213}
\definecolor{succBlue}{HTML}{4C78A8}
\definecolor{failOrange}{HTML}{F58518}
\definecolor{driftPurple}{HTML}{7A5195}
\definecolor{gridGray}{HTML}{D9D9D9}
\definecolor{axisGray}{HTML}{666666}
\colorlet{citecolor}{mydarkblue}
\definecolor{RubineRed}{cmyk}{0,1,0.13,0}

\pgfplotsset{
  failureplot/.style={
    width=\linewidth,
    height=5.35cm,
    ybar,
    axis line style={draw=axisGray, line width=0.8pt},
    tick style={draw=axisGray},
    ymajorgrids=true,
    grid style={dashed, draw=gridGray},
    xtick=data,
    xticklabel style={rotate=22, anchor=east, font=\scriptsize},
    yticklabel style={font=\scriptsize},
    ylabel style={font=\small},
    tick label style={font=\scriptsize},
    enlarge x limits=0.16,
    nodes near coords style={font=\scriptsize},
    legend style={
      draw=none,
      font=\scriptsize,
      /tikz/every even column/.append style={column sep=0.5cm}
    },
    symbolic x coords={EvoCUA-8B,Qwen3-VL-8B,gpt-oss-20b,AgentTrek-32B},
  }
}

\tcbset{
  promptbox/.style={
    enhanced,
    breakable,
    colback=black!2,
    colframe=black!20,
    colbacktitle=black!25,
    coltitle=white,
    boxrule=0.5pt,
    arc=1pt,
    boxsep=0pt,
    left=4pt,
    right=4pt,
    top=4pt,
    bottom=4pt,
    toptitle=2pt,
    bottomtitle=2pt,
    fonttitle=\bfseries\small,
    before skip=6pt,
    after skip=6pt,
  }
}

\DefineVerbatimEnvironment{PromptText}{Verbatim}{
  fontsize=\footnotesize,
  baselinestretch=0.95,
  breaklines=true,
  breaksymbolleft={},
  breaksymbolright={},
  breakanywhere=false,
  xleftmargin=0pt,
  xrightmargin=0pt
}

\newcommand{\huggingface}{\raisebox{-1.5pt}{\includegraphics[height=1.05em]{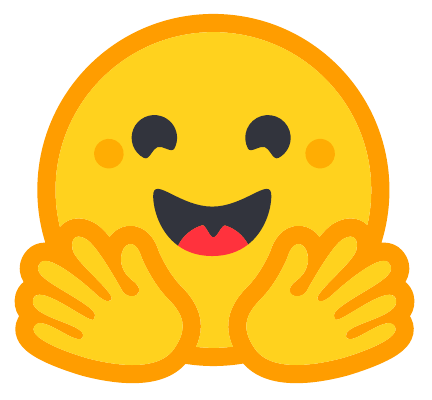}}\xspace}
\newcommand{\github}{\raisebox{-1.5pt}{\includegraphics[height=1.05em]{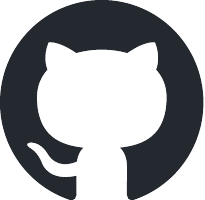}}\xspace}

\begin{document}

\title{Step-level Optimization for Efficient Computer-use Agents}

\author{
\textbf{Jinbiao Wei}$\Yale$\thanks{Equal contributions. Correspondence to: Jinbiao Wei (jinbiao.wei@yale.edu), Yilun Zhao (yilun.zhao@yale.edu).} \quad
\textbf{Kangqi Ni}$\UNC$\footnotemark[1] \quad
\textbf{Yilun Zhao}$\Yale$ \quad
\textbf{Guo Gan}$\Yale$ \quad
\textbf{Arman Cohan}$\Yale$ \\ [7pt]
$\Yale$Yale NLP Lab \quad $\UNC$University of North Carolina at Chapel Hill 
}

\maketitle
\thispagestyle{fancy}
\fancyhead{}
\lhead{%
    \includegraphics[height=1.1cm]{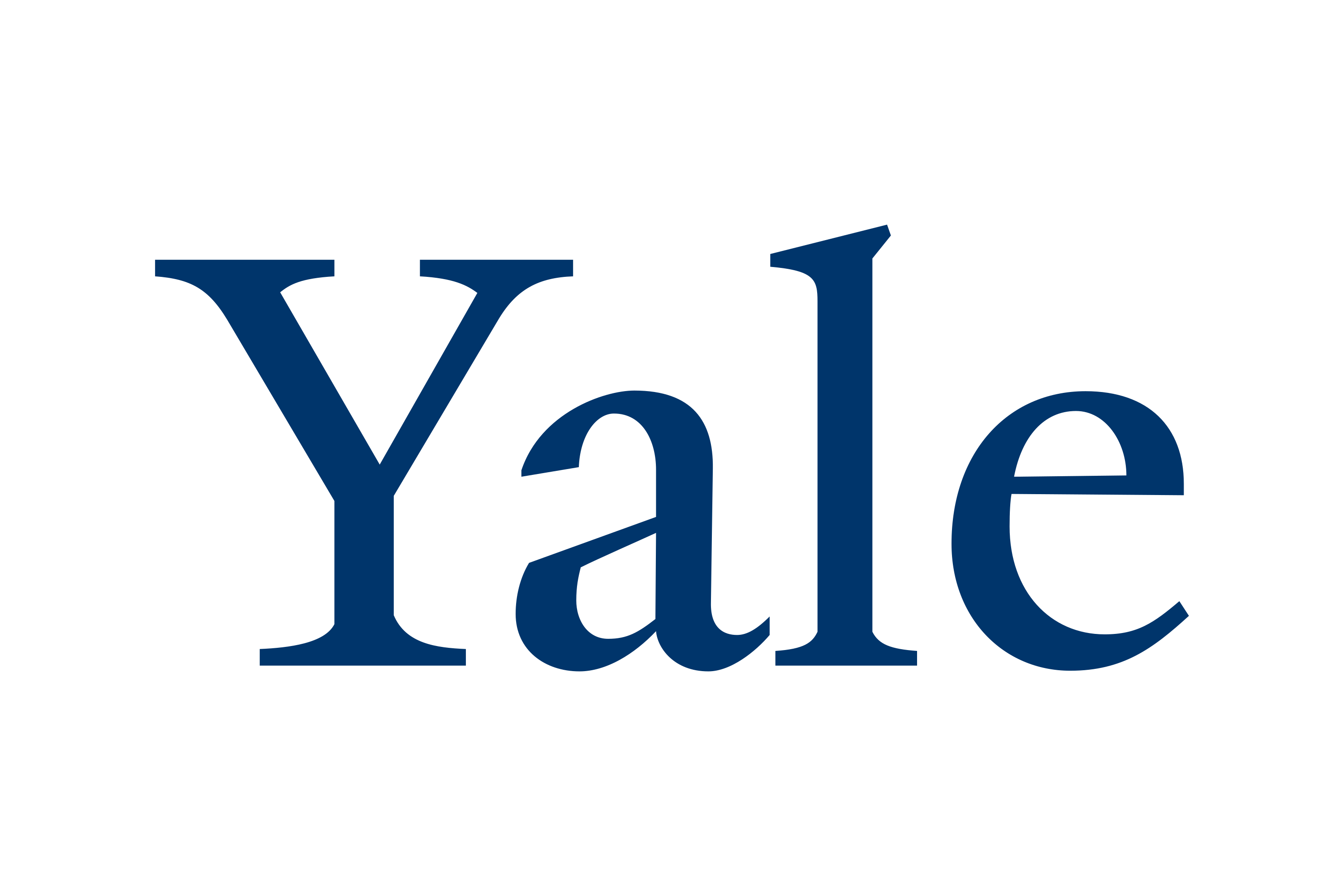}\hspace{0.1cm}%
    \raisebox{0.2cm}{\includegraphics[height=0.6cm]{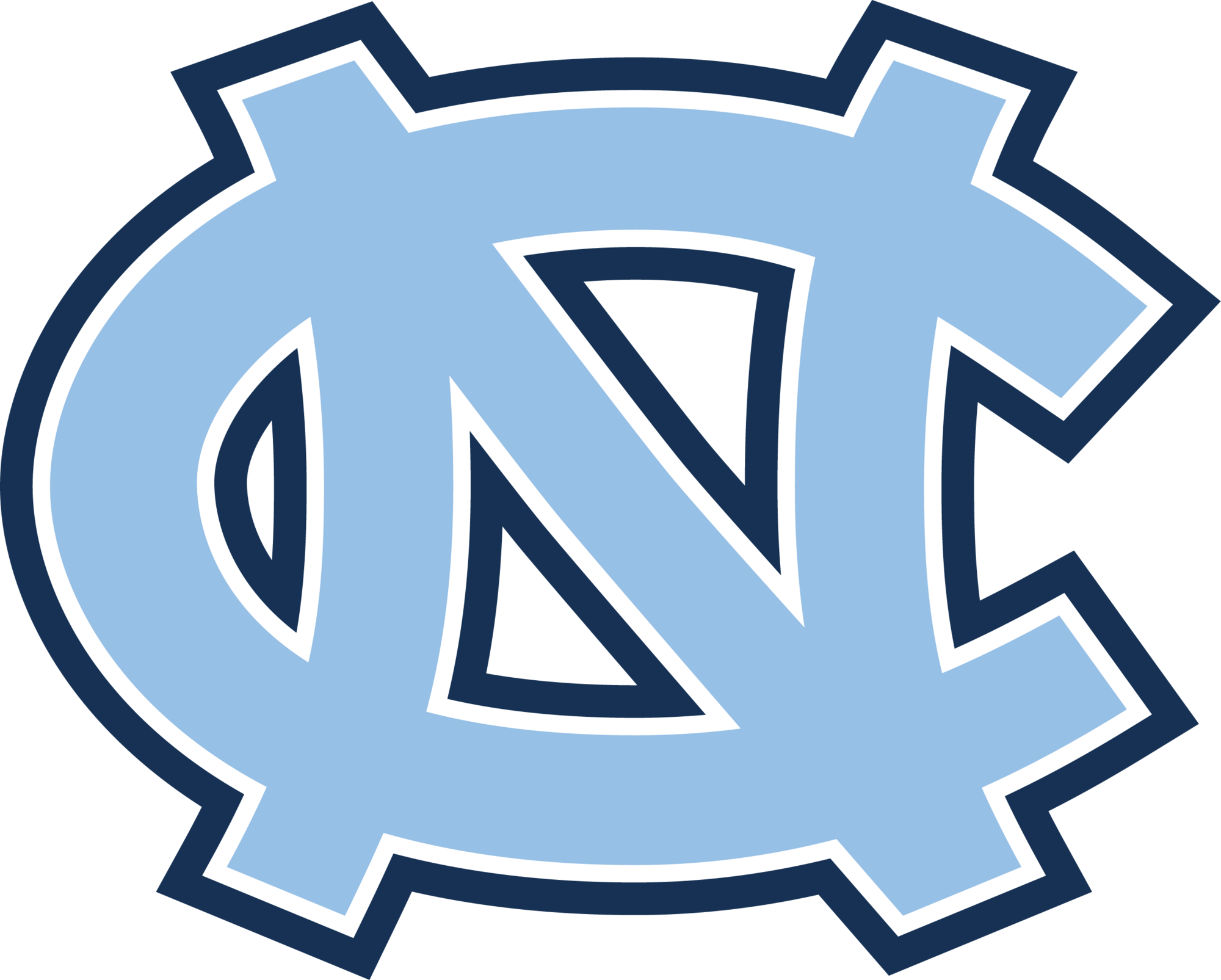}}\hspace{0.3cm}%
}

\fancyfoot[C]{\thepage}
\renewcommand{\headrulewidth}{0pt}
\setlength{\headheight}{12pt}
\addtolength{\topmargin}{0pt}
\setlength{\headsep}{3mm}

\vspace{-1.0em}

\pagestyle{plain}

\begin{abstract}
Computer-use agents provide a promising path toward general software automation because they can interact directly with arbitrary graphical user interfaces instead of relying on brittle, application-specific integrations. Despite recent advances in benchmark performance, strong computer-use agents remain expensive and slow in practice, since most systems invoke large multimodal models at nearly every interaction step. We argue that this uniform allocation of compute is fundamentally inefficient for long-horizon GUI tasks. Such trajectories are highly heterogeneous: many steps are routine and can be handled reliably by smaller, cheaper policies, while errors tend to concentrate at a relatively small number of high-risk moments. Across computer-use benchmarks, these failures repeatedly take two forms: progress stalls, where the agent loops, repeats ineffective actions, or fails to make meaningful progress, and silent semantic drift, where the agent continues taking locally plausible actions after already deviating from the user’s true goal. To address this inefficiency, we propose an event-driven, step-level cascade for computer-use agents that runs a small policy by default and escalates to a stronger model only when lightweight learned monitors detect elevated risk. Our framework combines two complementary signals: a Stuck Monitor that detects degraded progress from recent reasoning–action history and triggers recovery, and a Milestone Monitor that identifies semantically meaningful checkpoints where sparse verification is most informative for catching drift. This design turns always-on frontier-model inference into adaptive, on-demand compute allocation over the course of an evolving interaction. The framework is modular and deployment-oriented: it can be layered on top of existing computer-use agents without changing the underlying agent architecture or retraining the large model. Experiments on OSWorld and WebArena show that the proposed cascade consistently improves the performance–efficiency frontier, recovering much of the success rate of always-large policies while substantially reducing large-model usage, latency, and monetary cost. 
\end{abstract}

\begin{center}
\begin{tabular}{cl@{\hspace{5em}}cl}
\huggingface & \href{https://huggingface.co/collections/yale-nlp/stepwise}{\data} &
\github & \href{https://github.com/yale-nlp/StepWise}{\data}
\end{tabular}
\end{center}
\vspace{5pt}

\begin{figure*}[h]
  \centering
  \includegraphics[width=\textwidth]{figures/pipeline.png}
  \caption{
Overview of the proposed event-driven, step-level cascade for computer-use agents. A small policy acts by default while two lightweight monitors analyze the trajectory: a Stuck Event detector triggers escalation when recent steps show repeated actions or lack of progress, and a Milestone Event detector selects semantically meaningful checkpoints for sparse verification by a stronger model, which returns control to the small policy if the check passes and otherwise hands control to the large policy.
}
  \label{fig:pipeline}
\end{figure*}
\section{Introduction}

Computer-use agents offer an attractive path to automation: instead of building brittle, per-application integrations, an agent can interact directly with GUI interfaces and execute tasks across diverse applications~\citep{agasheagent, nguyen2025gui,agashe2025agent,song2025coact}. This makes it possible to execute end-to-end workflows on arbitrary software, including long-tail internal tools. Yet despite rapid gains in benchmark success rates~\citep{deng2023mind2webgeneralistagentweb, rawles2025androidworlddynamicbenchmarkingenvironment, xie2024osworldbenchmarking,zhou2023webarena,bonatti2024windowsagentarenaevaluating} enabled by increasingly effective training strategies~\cite{qin2025ui,wei2026anchor,gan2026android}, high-performing computer-use agents remain expensive and slow at inference time. Recent evaluations on realistic multi-step computer-use workloads report trajectories of 20--35 interaction steps, hundreds of thousands of tokens, and wall-clock runtimes in the tens of minutes, with per-task inference costs that can exceed one US dollar when large frontier models are used at every step~\citep{dai2025scuba}. Such latency and unit economics are difficult to justify in production settings, where throughput and cost matter just as much as raw success in real-world deployment.

A key inefficiency is that most agent frameworks allocate frontier-model compute uniformly: each interaction invokes the same large multimodal model to perceive the screen, maintain the plan, and select the next action. However, GUI trajectories are not uniformly difficult. Many steps are routine and can be executed reliably by much smaller models once local context is established. Conversely, failures tend to concentrate in a small number of high-risk moments and we classify them into two recurring modes: (i) \emph{progress stalls}, where the agent loops or repeatedly takes equivalent actions without changing the state; and (ii) \emph{silent semantic drift}, where the agent continues to act plausibly but has already deviated from the user’s intent (e.g., navigating to the wrong page or selecting an unintended setting), making the remainder of the trajectory locally consistent while dooming the final outcome. This heterogeneity suggests a different allocation strategy: treat large-model compute as an on-demand resource, triggered only when the trajectory exhibits evidence of elevated risk.

Building on this observation, we introduce an \emph{event-driven, step-level cascade} for computer-use agents. The system runs a small, inexpensive GUI policy by default and escalates to a stronger model only when a lightweight controller predicts degradation. Concretely, the controller comprises two learned monitors trained on short windows of recent interaction history. The \textbf{Stuck Monitor} predicts whether the agent is failing to make progress given the recent steps and the current action/reasoning; when it fires, the system escalates to a stronger model to recover and resume productive exploration. The \textbf{Milestone Monitor} predicts whether the current step completes a semantically meaningful checkpoint in the task. Milestones provide a natural unit for sparse oversight: verifying every step is costly, while verifying only at the end is often too late to correct semantic drift. When the Milestone Monitor fires, we trigger a targeted verification call to a stronger model.

We instantiate this approach with a deployment-oriented training and control pipeline. We first collect interaction traces from small agents across diverse tasks and interfaces. A stronger LLM (e.g., Claude Sonnet 4.5, GPT-5-series models, and Kimi K2.5) then provides supervision over short step windows, labeling segments that reflect healthy progress versus degraded execution (e.g., stuck behavior) and identifying steps that complete milestones. Using these labels, we train a compact classifier (i.e., a BERT-family encoder~\citep{warner2025smarter}) that consumes a rolling summary of recent interactions to produce calibrated risk scores. At runtime, these scores drive a stable policy with hysteresis and bounded recovery budgets: escalation is triggered when degradation persists, and the system escalates by switching control to a stronger model. Our framework is intentionally plug-and-play: it can be layered on top of existing computer-use agents without modifying the base architecture, retraining the large model, or designing task-specific heuristics. The monitors are trained from logged trajectories with simple labels (stuck vs.\ non-stuck; milestone-completed vs.\ not), and the cascade exposes clear operating points through thresholds that trade off cost against success. In this sense, the method acts as a deployment-time controller for computer-use agents, converting always-on frontier inference into event-driven escalation.

We evaluate the step-level cascade on interactive multi-step GUI benchmarks and find that it substantially improves the cost--quality trade-off. Compared to an always-large agent, our cascade achieves comparable task success while sharply reducing large-model usage, yielding meaningful savings in wall-clock time and dollar cost. Ablation studies further show that the gains come from combining two complementary routing signals: stuck detection improves recovery from local failure modes, while milestone-triggered verification helps catch silent semantic drift. We also find the event-driven framework is more effective and more efficient than fixed-interval checking, and that lightweight monitors provide sufficiently accurate signals to support reliable step-level control.

We summarize our contributions as follows:
\begin{enumerate}[leftmargin=*, itemsep=2pt]
    \item \textbf{A systematic study of inference efficiency in computer-use agents.}
    We characterize where inference cost comes from in realistic multi-step computer-use trajectories and show that failures and ``hard'' decisions concentrate in a small subset of steps, motivating adaptive compute allocation.

    \item \textbf{An event-driven framework for efficient inference.}
    We propose a step-level framework that runs a small model by default and switches to a stronger model when lightweight monitors detect elevated risk. The framework is modular and can be trained from logged trajectories without modifying the underlying agent.

    \item \textbf{Comparable success with major savings in both cost and time.}
    On realistic computer-use benchmarks, our framework achieves comparable task success to always-large agents while reducing inference cost by up to 74.6\% and latency by up to 45.8\%.
\end{enumerate}

\section{Related Work}

\paragraph{Routing and cascading for cost--quality trade-offs.}
A large body of work studies \emph{model selection} to balance quality against latency and monetary cost. 
In the standard LLM setting, \emph{routing} chooses one model per query, while \emph{cascading} escalates to stronger models only when needed, relying on accurate quality estimators or stopping criteria, as in FrugalGPT, RouteLLM, and Hybrid LLM~\citep{chen2023frugalgpt,ong2024routellm,ding2024hybrid}.
Beyond query-level selection, recent work extends routing to \emph{agentic} systems, where the controller must also decide collaboration structure (e.g., roles, interaction patterns) and route among different models, as in agent routing frameworks such as MasRouter~\citep{yue2025masrouter}, xRouter~\citep{qian2025xroutertrainingcostawarellms}, and EvoRoute~\citep{zhang2026evorouteexperiencedrivenselfroutingllm}.
However, these formulations still differ from computer-use agents: model selection becomes a far more fine-grained, state-dependent control problem. The decision of when to escalate is often driven by subtle, step-specific cues in the evolving UI state and interaction history, making computer-use agent model selection fundamentally more nuanced than query-level routing and motivating a dedicated framework for step-wise selection and switching.

\paragraph{Efficiency and practicality of computer-use agents.}
While early work on computer-use agents primarily emphasized task success, recent studies highlight that \emph{latency and cost} are often the bottleneck for real deployment.
OSWorld-Human~\citep{abhyankar2025osworld} provides a focused efficiency benchmark and temporal analysis. SCUBA~\citep{dai2025scuba} similarly emphasizes enterprise realism and reports both time and monetary cost alongside success, showing that demonstration augmentation can improve success while also reducing time and cost.
Complementary to benchmarking, Fara-7B~\citep{awadallah2025fara} targets efficiency at the model level by training a small (7B-level), native computer-use agent via scalable data generation, and Ferret-UI Lite~\citep{yang2025ferret} distills lessons for building small on-device agents with curated mixtures, inference-time strategies, and RL refinements. Despite this progress, existing work largely focused either on evaluation or on training more efficient backbone models, rather than on \emph{inference-time, step-wise model optimization} for computer-use agents. We fill this gap with a dedicated framework for step-level optimization and stable switching within long-horizon GUI interaction.

\section{Failure Modes and Motivation for Step-Level Optimization}
\label{sec:problem}

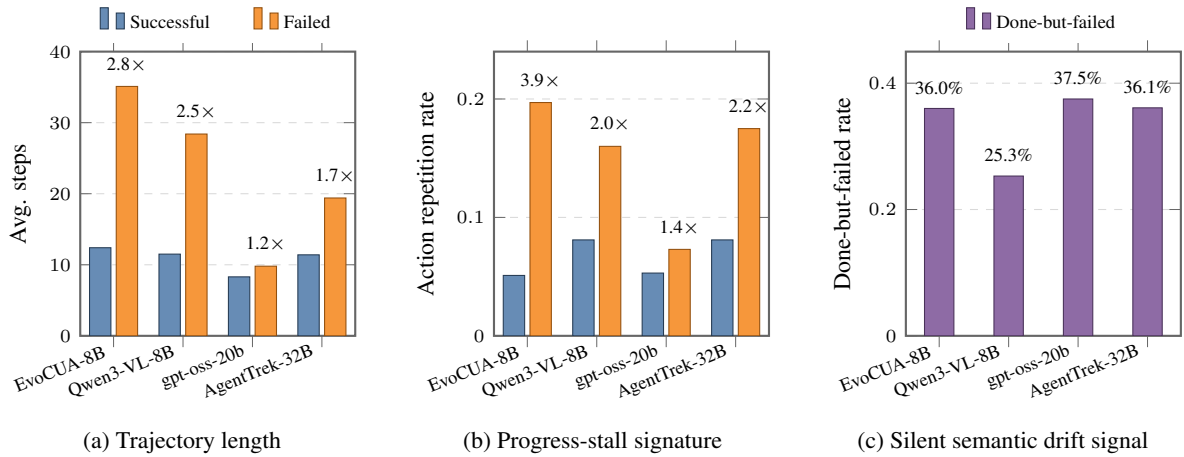
\begin{figure*}[t]
\centering

\begin{subfigure}[t]{0.325\textwidth}
\centering
\begin{tikzpicture}
\begin{axis}[
    failureplot,
    bar width=8pt,
    ymin=0,
    ymax=40,
    ylabel={Avg. steps},
    legend style={
        at={(0.5,1.03)},
        anchor=south,
        legend columns=2,
        draw=none,
        font=\scriptsize
    },
]
\addplot+[
    fill=succBlue!85,
    draw=succBlue!50!black
] coordinates {
    (EvoCUA-8B,12.4)
    (Qwen3-VL-8B,11.5)
    (gpt-oss-20b,8.3)
    (AgentTrek-32B,11.4)
};

\addplot+[
    fill=failOrange!85,
    draw=failOrange!60!black,
    nodes near coords,
    every node near coord/.append style={yshift=2pt, text=black},
    point meta=explicit symbolic
] coordinates {
    (EvoCUA-8B,35.1) [2.8$\times$]
    (Qwen3-VL-8B,28.4) [2.5$\times$]
    (gpt-oss-20b, 9.8) [1.2$\times$]
    (AgentTrek-32B,19.4) [1.7$\times$]
};
\legend{Successful, Failed}
\end{axis}
\end{tikzpicture}
\caption{Trajectory length}
\end{subfigure}
\hfill
\begin{subfigure}[t]{0.325\textwidth}
\centering
\begin{tikzpicture}
\begin{axis}[
    failureplot,
    bar width=8pt,
    ymin=0,
    ymax=0.24,
    ylabel={Action repetition rate},
    yticklabel style={
      /pgf/number format/fixed,
      /pgf/number format/precision=2,
      font=\scriptsize
    },
]
\addplot+[
    fill=succBlue!85,
    draw=succBlue!50!black
] coordinates {
    (EvoCUA-8B,0.051)
    (Qwen3-VL-8B,0.081)
    (gpt-oss-20b,0.053)
    (AgentTrek-32B,0.081)
};

\addplot+[
    fill=failOrange!85,
    draw=failOrange!60!black,
    nodes near coords,
    every node near coord/.append style={yshift=2pt, text=black},
    point meta=explicit symbolic
] coordinates {
    (EvoCUA-8B,0.197) [3.9$\times$]
    (Qwen3-VL-8B,0.160) [2.0$\times$]
    (gpt-oss-20b,0.073) [1.4$\times$]
    (AgentTrek-32B,0.175) [2.2$\times$]
};
\end{axis}
\end{tikzpicture}
\caption{Progress-stall signature}
\end{subfigure}
\hfill
\begin{subfigure}[t]{0.325\textwidth}
\centering
\begin{tikzpicture}
\begin{axis}[
    failureplot,
    bar width=11pt,
    ymin=0,
    ymax=0.45,
    ylabel={Done-but-failed rate},
    legend style={
        at={(0.5,1.03)},
        anchor=south,
        legend columns=1,
        draw=none,
        font=\scriptsize
    },
    yticklabel style={
      /pgf/number format/fixed,
      /pgf/number format/precision=1,
      font=\scriptsize
    },
]
\addplot+[
    fill=driftPurple!85,
    draw=driftPurple!60!black,
    nodes near coords,
    every node near coord/.append style={font=\scriptsize, yshift=2pt, text=black},
    point meta=explicit symbolic
] coordinates {
    (EvoCUA-8B,0.360) [36.0\%]
    (Qwen3-VL-8B,0.253) [25.3\%]
    (gpt-oss-20b,0.375) [37.5\%]
    (AgentTrek-32B,0.361) [36.1\%]
};
\legend{Done-but-failed}
\end{axis}
\end{tikzpicture}
\caption{Silent semantic drift signal}
\end{subfigure}

\caption{
Quantitative failure signatures of small-policy trajectories.
(a) Failed episodes are substantially longer than successful ones, indicating prolonged unproductive execution.
(b) Failed episodes exhibit markedly higher action repetition rates, consistent with progress stalls.
(c) A non-trivial fraction of failures terminate with a model-issued \texttt{done}, revealing silent semantic drift.
}
\label{fig:failure-signatures}
\end{figure*}

\paragraph{Failure modes in long-horizon GUI execution.}
Modern computer-use agents operate in long closed-loop trajectories: each step both consumes compute and irreversibly changes the interface state. In this setting, failures are rarely uniformly distributed across the trajectory. Instead, they tend to concentrate in a small number of high-risk moments. We focus on two recurring modes that are damaging in practice: (i) \emph{progress stalls}, where the agent loops, repeats equivalent actions, or fails to make meaningful change for the task; and (ii) \emph{silent semantic drift}, where actions remain locally plausible but the agent has already deviated from the user intent, so the remainder of the trajectory is coherent yet ultimately doomed.
\autoref{fig:failure-signatures} makes this pattern concrete. Across multiple small default policies, failed episodes are consistently much longer than successful ones---for example, EvoCUA-8B failures average 2.8$\times$ more steps than successes, and Qwen3-VL-8B shows a similarly large 2.5$\times$ gap. This suggests that many failures are not caused by a single bad decision, but by extended stuck regions in which the agent keeps acting without making real progress. Consistent with this interpretation, failed trajectories also exhibit substantially higher action repetition rates, matching progress stalls such as repeated clicks, oscillations, or ineffective retries. At the same time, these failures are not purely looping behavior: a sizable fraction of failed episodes still terminate with a model-issued \texttt{done}, indicating that some trajectories remain locally coherent while drifting away from the true task objective. This pattern is consistent with silent semantic drift. Together, these two failure modes motivate complementary escalation signals.

\paragraph{Why query-level cascades are not enough.}
Standard LLM cascading typically makes a single routing decision \emph{per query}: the system estimates difficulty, or evaluates a candidate answer, and then decides whether to escalate to a stronger model. Computer-use agents are different. Because they interact with an evolving environment over many steps, it is both possible and desirable to perform cascading at the \emph{step level}. However, this is more complex than standard query-level routing. Instead of judging a completed answer, the controller must assess whether an \emph{ongoing interaction process} is still on track. Intermediate actions are weak proxies for correctness: a step may look locally reasonable while already inducing semantic drift, and its consequences may only become apparent several steps later. As a result, naively verifying every step with a stronger model would erase much of the efficiency benefit of cascading. Moreover, difficulty is \emph{state-dependent} and can change abruptly within a single task; errors \emph{compound} over time and may remain latent for many steps; and routing must remain stable enough (avoid thrashing between models) while still enabling timely recovery. Step-level model selection is therefore necessary for computer-use agents, but it is also fundamentally harder than static selection because the system must decide during an evolving interaction.

\section{Methodology}
\label{sec:method}

We propose an \emph{event-driven, step-level cascade} that runs a small GUI policy by default and allocates expensive large-model compute only when there is evidence of elevated risk.~\autoref{fig:pipeline} illustrates the overall architecture of our framework. The controller relies on two lightweight learned monitors that trigger \emph{events}: a \emph{Stuck Monitor} that detects progress degradation and requests recovery, and a \emph{Milestone Monitor} that identifies semantically meaningful checkpoints where it is most informative to verify intent and progress.

\paragraph{Setup and notation.}
A GUI task is an episode of length at most $T$ with observations $o_t$ (e.g., screenshot and optional accessibility/DOM signals) and interaction history $h_t$. We are given two policies/models with different cost profiles, $M=\{\pi_{\text{small}},\pi_{\text{large}}\}$. At step $t$, the selected policy $m_t\in M$ produces both a textual \emph{step rationale} $r_t$ (brief reasoning/plan for the next move) and an executable action $a_t$ (e.g., click/type/hotkey with arguments). The action is executed in the environment and the episode terminates on success or after $T$ steps. Our goal is to improve the success--efficiency trade-off by selecting $m_t$ online.

\paragraph{Compact window of model outputs.}
At each step $t$ we build a compact window of the last $K$ model outputs:
\[
w_t = \big\{(r_{t-K+1}, a_{t-K+1}), \ldots, (r_t, a_t)\big\}.
\]
This window is cheap to construct because it is directly available from the agent’s own generation logs, and it captures local behavioral patterns such as repeated rationales, oscillating plans, or action loops.

\paragraph{Two monitors with different inputs.}
We instantiate both monitors with lightweight ModernBERT~\citep{warner2025smarter} encoders with distinct conditioning:
\[
p^{\text{stuck}}_t = S_\phi(w_t), \qquad
p^{\text{mile}}_t = M_\psi(u, w_t),
\]
where $u$ is the task description. The \textbf{Stuck Monitor} $S_\phi$ only uses the local window because stuckness is primarily a \emph{behavioral} phenomenon (e.g., repeated plans/actions, cycling). In contrast, the \textbf{Milestone Monitor} $M_\psi$ additionally conditions on the task description because milestones are \emph{goal-dependent}: the same local behavior may or may not constitute meaningful progress depending on $u$. To keep step-level routing efficient, we intentionally restrict these always-on monitors to textual reasoning--action traces rather than raw screenshots or DOM deltas. This choice makes the controller lightweight, model-agnostic, and inexpensive enough to run at every step.

\paragraph{Event-driven control via stuck detection.}
The controller uses the Stuck Monitor to decide when to allocate large-model compute for
recovery. After each interaction step $t$, the monitor outputs a stuck score
$p^{\text{stuck}}_t = S_{\phi}(w_t)$ based on the recent window $w_t$.
If this score exceeds a threshold $\theta_s$, we treat the trajectory as stalled and
escalate to the stronger policy for the next decision:
\[
E_{t+1} = \mathbb{I}\!\left[p^{\text{stuck}}_t \ge \theta_s\right].
\]
At the next step, we route action generation according to this escalation indicator:
\[
m_t =
\begin{cases}
\pi_{\text{large}}, & \text{if } E_t = 1, \\
\pi_{\text{small}}, & \text{otherwise.}
\end{cases}
\]
To ensure a seamless handoff, when we route to $\pi_{\text{large}}$ we inject the small model's
recent history by re-serializing it into the
large model’s response format, so the large model can directly continue from the current state
with full local context.

\paragraph{Milestone detection and verification.}
Milestones serve a different purpose: they identify sparse checkpoints where we can cheaply audit whether the agent’s intermediate outcome still matches the user’s intent. Let $\tau_t$ denote the most recent verified milestone index (with $\tau_t=0$ at the start). When $p^{\text{mile}}_t \ge \theta_m$, we form a \emph{milestone packet} that summarizes the transition from $\tau_t$ to $t$:
(i) the task instruction $u$,
(ii) the rationale--action trace $\{(r_{\tau_t+1},a_{\tau_t+1}),\ldots,(r_t,a_t)\}$, and
(iii) visual evidence consisting of a screenshot at the current state and the screenshot saved at step $\tau_t$.
A stronger verifier model then answers two questions:
\textbf{(1)} \emph{Progress validity}: is the local plan/trajectory from $\tau_t$ to $t$ plausibly advancing toward the goal?
\textbf{(2)} \emph{Intent consistency}: does the UI state at $t$ reflect the intended outcome implied by $u$ and the agent’s rationales (i.e., no semantic drift)?
If either check fails, we treat this as elevated risk and escalate.
If verification passes, we \emph{commit} the milestone by setting $\tau_{t+1}\!\leftarrow t$ and saving the screenshot at $t$ for the next checkpoint. This makes verification both \emph{sparse} (only at predicted milestones) and \emph{localized} (only the segment since the last milestone), while still providing strong global oversight.

\paragraph{Training data via LLM supervision.}
We collect trajectories by running $\pi_{\text{small}}$ across diverse tasks and interfaces and extract overlapping windows $\{w_t\}$. A stronger LLM labels each window with (1) stuck vs.\ non-stuck using \emph{only} $w_t$, and (2) milestone-completed vs.\ not using $(u,w_t)$. We further reduce annotation noise by running LLM multiple times per trajectory and keeping only high-consensus labels, while discarding low-agreement cases. We then train $S_\phi$ and $M_\psi$ using cross-entropy with class balancing. In deployment, thresholds $(\theta_s,\theta_m)$ are selected to trade off success against cost/latency. Additional ModernBERT fine-tuning details are provided in Appendix~\ref{app:training-details}, and the full labeling and verification prompts are listed in Appendix~\ref{app:prompts}.

\section{Experiments}
\subsection{Experimental Setup}

\paragraph{Benchmarks.}
We evaluate our step-level cascading framework on two widely used computer-use agent benchmarks: \textbf{OSWorld}~\citep{xie2024osworldbenchmarking} and \textbf{WebArena}~\citep{zhou2023webarena}. \textbf{OSWorld} focuses on desktop-style computer-use tasks across applications such as browsers, office tools, coding environments, and system utilities, making it a representative testbed for long-horizon GUI interaction in general-purpose operating-system settings. \textbf{WebArena}, in contrast, evaluates web-based agents on realistic online tasks spanning multiple websites and domains. Together, these benchmarks cover two complementary settings for computer-use agents---desktop interaction and web navigation---and provide a broad view of how cascading behaves under different types of environments. Additional benchmark-specific environment and evaluation details are provided in Appendix~\ref{app:benchmarks}.

\paragraph{Models.}
We evaluate both standalone models and cascaded model pairs. On \textbf{OSWorld}, the small-model pool consists of \textbf{Qwen3-VL-8B}~\citep{yang2025qwen3technicalreport} and \textbf{EvoCUA-8B}~\citep{xue2026evocuaevolvingcomputeruse}, while the large-model pool includes \textbf{Claude Sonnet 4.5}~\citep{anthropic_claude_sonnet_45_2025} and \textbf{Kimi K2.5}~\citep{moonshot_kimi_k25_2026}. On \textbf{WebArena}, we use \textbf{gpt-oss-20b}~\citep{openai_gpt_oss_20b_2025} and \textbf{AgentTrek 32B}~\citep{xu2024agenttrek} as small models, and \textbf{GPT-5 mini}~\citep{openai_gpt_5_mini_2025} and \textbf{GPT-5.2}~\citep{openai_gpt_5_2_2025} as large models. We then construct cascaded settings by pairing each small model with a stronger large model. This design lets us test whether a small default policy, augmented with selective escalation to a stronger model, can achieve a better performance--efficiency trade-off than either model alone.

\paragraph{Evaluation.}
For both benchmarks, we evaluate every \emph{single-model baseline} and every \emph{cascaded configuration} under the same task setting. In cascaded runs, the smaller model acts as the default policy, and the stronger model is called only when the controller decides to escalate. This setup allows us to directly measure whether step-level routing can preserve the performance benefits of stronger models while reducing inference cost.
We evaluate both \textbf{effectiveness} and \textbf{efficiency} on \textbf{OSWorld} and \textbf{WebArena}. Specifically, for both benchmarks we report overall \textbf{task success rate} and \textbf{inference cost}. We also report efficiency-related statistics for both benchmarks, including latency, cost, and behavior in cascaded settings. For models derived from open-weight base models, we report reference cost estimates using OpenRouter~\citep{openrouter_pricing} pricing. For fine-tuned models, these estimates are based on the corresponding open-weight base model available. Reported latency is measured from our local deployment using 2× H100 GPUs. Taken together, these metrics provide a unified view of the performance--efficiency trade-off and allow us to assess when step-level cascading offers advantages over standalone models.

\begin{table*}[t]
\centering
\small
\setlength{\tabcolsep}{4pt}
\resizebox{\textwidth}{!}{%
\begin{tabular}{llccccccc}
\toprule
& & \multicolumn{3}{c}{\textbf{Efficiency}} & \multicolumn{4}{c}{\textbf{Cascading Statistics}} \\
\cmidrule(lr){3-5} \cmidrule(lr){6-9}
\textbf{Small} & \textbf{Large} & \textbf{Lat./Req.} & \textbf{Cost/Task} & \textbf{Acc.} & \textbf{Avg Step} & \textbf{Switched} & \textbf{A1 Share} & \textbf{A2 Share} \\
\midrule
EvoCUA-8B  & --                & 2.6s         & \$0.022 & 43.3\% & 30.9 & --           & --      & -- \\
Qwen3-VL-8B   & --                & 3.9s         & \$0.018 & 30.8\% & 23.6 & --           & --      & -- \\
--         & Claude Sonnet 4.5 & 6.4s        & \$0.881 & 58.1\% & 25.4 & --           & --      & -- \\ 
--         & Kimi K2.5          & 8.3s        & \$0.132 & 60.1\% & 22.4 & --           & --      & -- \\
\midrule
EvoCUA-8B  & Claude Sonnet 4.5 & 4.1s & \$0.224 & 55.4\% & 26.2 & 168 (46.8\%) & 60.6\% & 39.4\% \\
EvoCUA-8B  & Kimi K2.5          & 4.5s & \$0.051 & 58.2\% & 25.2 & 173 (48.2\%) & 59.5\% & 40.5\% \\
Qwen3-VL-8B   & Claude Sonnet 4.5 & 5.2s           & \$0.423     & 54.3\%   & 24.2    & 234 (65.2\%)     & 38.4\% & 61.6\% \\
Qwen3-VL-8B   & Kimi K2.5          & 6.5s           & \$0.078 & 59.3\% & 23.7 & 240 (66.9\%) & 37.4\% & 62.6\% \\
\bottomrule
\end{tabular}
}
\caption{OSWorld efficiency, cost, and cascading statistics. 
``Switched'' denotes the number of tasks that invoked the large model at least once. 
``A1'' and ``A2'' denote the fraction of executed steps assigned to the small and large model, respectively.}
\label{tab:osworld-efficiency}
\end{table*}
\subsection{Main Results}
\autoref{tab:osworld-efficiency} and~\autoref{tab:webarena-efficiency} show that step-level cascading consistently improves the performance--efficiency frontier. Across both benchmarks, the same pattern holds: standalone small models are cheap but much weaker, while always-large policies deliver the highest raw success at substantially higher monetary and latency cost. Our cascaded policies achieve the most favorable middle ground, recovering most of the performance of always-large inference without paying to invoke the stronger model at every step.

On OSWorld, cascading comes close to the strongest always-large policy while remaining much cheaper. For example, Qwen3-VL-8B + Kimi K2.5 reaches 59.3\% success, very close to the 60.1\% of standalone Kimi K2.5, but at much lower cost. EvoCUA-8B + Kimi K2.5 is even more efficient, achieving 58.2\% success at only \$0.051 per task, reducing the cost by 61.4\%. Notably, both Kimi-based cascades already match or slightly exceed the standalone Claude Sonnet 4.5 baseline, showing that selective escalation is sufficient to recover most of the benefit of a strong model on long-horizon desktop tasks.

WebArena shows the same qualitative trend, though with a slightly larger gap to the strongest always-large policy. Standalone GPT-5.2 performs best overall, and both GPT-5.2-based cascades remain close to it while outperforming their standalone small-model counterparts. They also reduce cost and latency relative to always-large GPT-5.2, indicating that event-driven escalation remains effective even in shorter web trajectories.

A further pattern is that final success depends more on the choice of the \emph{large} model than on the exact choice of the \emph{small} model. For a fixed large model, swapping the small policy typically changes accuracy only modestly, whereas upgrading the large model produces a much larger gain. This suggests that the large model primarily determines the attainable performance ceiling, while the small model mainly serves as an efficient default policy that reduces how often expensive escalation is needed. In this sense, the small model mostly governs cost, and the large model mostly governs final accuracy.

\begin{table*}[t]
\centering
\small
\setlength{\tabcolsep}{4pt}
\resizebox{\textwidth}{!}{%
\begin{tabular}{llccccccc}
\toprule
& & \multicolumn{3}{c}{\textbf{Efficiency}} & \multicolumn{4}{c}{\textbf{Cascading Statistics}} \\
\cmidrule(lr){3-5} \cmidrule(lr){6-9}
\textbf{Small} & \textbf{Large} & \textbf{Lat./Req.} & \textbf{Cost/Task} & \textbf{Acc.} & \textbf{Avg Step} & \textbf{Switched} & \textbf{A1 Share} & \textbf{A2 Share} \\
\midrule
AgentTrek-32B & --            & 3.9s & \$0.005 & 30.3\% & 16.0 & -- & -- & -- \\
gpt-oss-20b   & --            & 6.9s & \$0.001 & 37.6\% & 8.1  & -- & -- & -- \\
--            & GPT-5 mini    & 11.1s & \$0.053 & 55.0\% & 13.8 & -- & -- & -- \\
--            & GPT-5.2       & 19.6s & \$0.335 & 60.1\% & 9.9  & -- & -- & -- \\
\midrule
gpt-oss-20b & GPT-5 mini & 8.5s & \$0.027 & 51.3\% & 12.1 & 568 (69.9\%) & 30.6\% & 69.4\% \\
gpt-oss-20b & GPT-5.2 & 12.2s & \$0.211 & 57.8\% & 10.3 & 578 (71.2\%) & 33.1\% & 66.9\% \\
AgentTrek-32B & GPT-5 mini & 9.4s & \$0.028 & 52.3\% & 8.2 & 562 (69.2\%) & 40.1\% & 59.9\% \\
AgentTrek-32B & GPT-5.2 & 13.4s & \$0.208 & 58.8\% & 12.0 & 593 (73.0\%) & 43.7\% & 56.3\% \\
\bottomrule
\end{tabular}
}
\caption{WebArena efficiency, cost, and cascading statistics. ``Switched'' denotes the number of tasks that invoked the large model at least once. ``A1'' and ``A2'' denote the fraction of executed steps assigned to the small and large model, respectively.}
\label{tab:webarena-efficiency}
\end{table*}

\section{Analysis}
\subsection{Component Ablation}

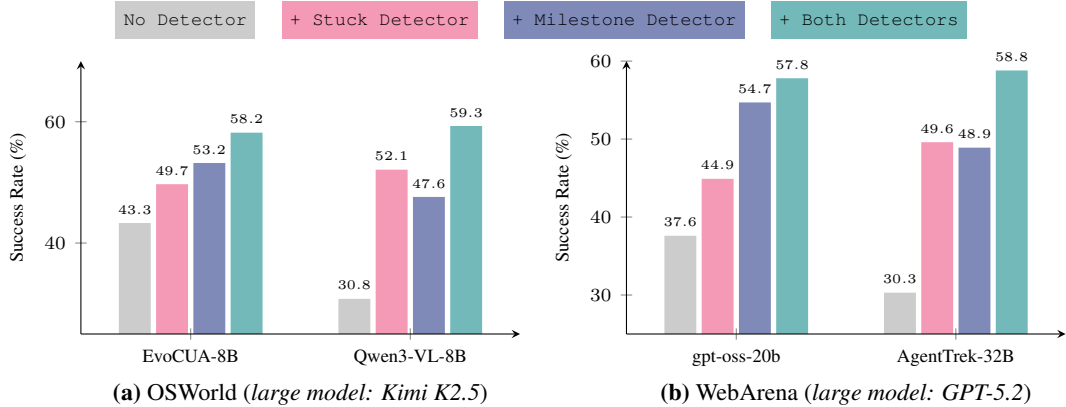
\begin{figure*}[htb]
    \centering
    \begin{tikzpicture}
    \colorlet{baseOnly}{black!20}
    \colorlet{stuckOnly}{RubineRed!50}
    \colorlet{mileOnly}{citecolor!50}
    \colorlet{bothDet}{teal!55}

    \begin{groupplot}[
        group style={
            group size=2 by 1,
            horizontal sep=14mm,
            group name=plots
        },
    ]

\nextgroupplot[
  width=0.46\textwidth,
  height=5.2cm,
  ybar,
  ymin=25, ymax=70,
  xmin=-0.5, xmax=1.5,
  bar width=4.2mm,
  enlarge x limits=0.22,
  font=\scriptsize,
  xticklabels={EvoCUA-8B, Qwen3-VL-8B},
  xtick=data,
  xlabel={\small \textbf{(a)} OSWorld (\textit{large model: Kimi K2.5})},
  xlabel near ticks,
  ylabel={Success Rate (\%)},
  ylabel style={yshift=0mm},
  axis lines={left},
  axis on top,
  set layers=standard,
  nodes near coords,
  every node near coord/.append style={
    /pgf/number format/precision=1,
    /pgf/number format/fixed zerofill,
    /pgf/number format/fixed,
    font=\fontsize{4.5pt}{5pt}\selectfont
  },
]

\addplot[ybar, draw=none, fill=baseOnly]
coordinates {(0,43.3) (1,30.8)};
\addplot[ybar, draw=none, fill=stuckOnly]
coordinates {(0,49.7) (1,52.1)};
\addplot[ybar, draw=none, fill=mileOnly]
coordinates {(0,53.2) (1,47.6)};
\addplot[ybar, draw=none, fill=bothDet]
coordinates {(0,58.2) (1,59.3)};

\nextgroupplot[
  width=0.46\textwidth,
  height=5.2cm,
  ybar,
  ymin=25, ymax=60,
  xmin=-0.5, xmax=1.5,
  bar width=4.2mm,
  enlarge x limits=0.22,
  font=\scriptsize,
  xticklabels={gpt-oss-20b, AgentTrek-32B},
  xtick=data,
  xlabel={\small \textbf{(b)} WebArena (\textit{large model: GPT-5.2})},
  xlabel near ticks,
  ylabel={Success Rate (\%)},
  ylabel style={yshift=0mm},
  axis lines={left},
  axis on top,
  set layers=standard,
  nodes near coords,
  every node near coord/.append style={
    /pgf/number format/precision=1,
    /pgf/number format/fixed zerofill,
    /pgf/number format/fixed,
    font=\fontsize{4.5pt}{5pt}\selectfont
  },
]

\addplot[ybar, draw=none, fill=baseOnly]
coordinates {(0,37.6) (1,30.3)};
\addplot[ybar, draw=none, fill=stuckOnly]
coordinates {(0,44.9) (1,49.6)};
\addplot[ybar, draw=none, fill=mileOnly]
coordinates {(0,54.7) (1,48.9)};
\addplot[ybar, draw=none, fill=bothDet]
coordinates {(0,57.8) (1,58.8)};

    \end{groupplot}

\node[
    matrix of nodes,
    draw=none,
    inner sep=0.5em,
    at={(current bounding box.south)},
    anchor=north,
    yshift=+5.65cm,
    xshift=+0.10cm,
    nodes={font=\scriptsize, inner sep=0em},
    column sep=0em,
]
{
    \colorbox{baseOnly}{\textcolor{black}{\strut \texttt{No Detector}}} &
    |[minimum width=1em]| {} &
    \colorbox{stuckOnly}{\textcolor{black}{\strut \texttt{+ Stuck Detector}}} &
    |[minimum width=1em]| {} &
    \colorbox{mileOnly}{\textcolor{black}{\strut \texttt{+ Milestone Detector}}} &
    |[minimum width=1em]| {} &
    \colorbox{bothDet}{\textcolor{black}{\strut \texttt{+ Both Detectors}}} \\
};

\end{tikzpicture}
\vspace{-1em}
\caption{Ablation results showing the contribution of different detectors to final accuracy. Each group compares four routing settings: no detector, only the stuck detector, only the milestone detector, and both detectors enabled.}
\label{fig:ablation_detectors}
\end{figure*}

\autoref{fig:ablation_detectors} isolates the contribution of the two routing signals by comparing four variants: no detector, stuck-only, milestone-only, and the full system with both detectors enabled. Across both OSWorld and WebArena, either detector alone improves over the no-detector baseline, while using both yields the best overall performance. This shows that the benefit of step-level cascading does not come from a single trigger, but from combining two complementary signals for escalation. A notable pattern is that the gain from enabling both detectors is consistently larger than the gain from either individual component, indicating that the two signals cover different parts of the failure space rather than providing redundant supervision.

The stuck detector mainly helps when the agent falls into local failure modes such as repetition, oscillation, or short progress loops. The milestone detector plays a different role: it selects semantically meaningful checkpoints where verification by the stronger model is most informative, helping catch cases where the trajectory appears locally reasonable but has already drifted from the task.

The ablation trends are largely consistent with the failure signatures in~\autoref{fig:failure-signatures}. Policies with stronger progress-stall behavior benefit more from stuck detection, while those with weaker repetition but more silent semantic drift benefit more from milestone verification. This helps explain why gpt-oss-20b improves less under stuck-only routing and more under milestone-only routing. Overall, the two signals are complementary: stuck detection targets local looping failures, whereas milestone verification is more effective for semantically off-track trajectories that still appear locally coherent to the weaker policy.

Still, the full system performs best overall, suggesting that the two detectors are complementary rather than redundant. Explicit stuck detection improves responsiveness to immediate local failures, while milestone detection provides higher-level oversight of meaningful progress. Combining both gives the strongest and most consistent gains across settings and across benchmarks.

\subsection{Event-Driven vs.\ Fixed-Interval Escalation}
We further compare the event-driven framework against a fixed-interval baseline that triggers verification at uniform step intervals, rather than at learned events. Specifically, the periodic baseline invokes the same large-model check every
$k$ steps, independent of the current trajectory state or semantic progress. We sweep $k \in \{3,5,7\}$ and report the best-performing periodic setting in~\autoref{tab:verification-policy-ablation}. We conduct this comparison on the strong cascade pair for each benchmark: \textit{EvoCUA-8B + Kimi K2.5} on OSWorld and \textit{AgentTrek-32B + GPT-5.2} on WebArena. This provides a strong test against a simple sparse-checking strategy that ignores task structure and verifies at uniformly spaced steps. Overall, periodic verification is consistently less efficient: across both benchmarks, even its best setting incurs substantially higher cost per task than the event-driven framework.
On OSWorld, the strongest periodic baseline remains competitive in accuracy, but only at a substantially higher cost than the full event-driven policy. Because OSWorld trajectories are relatively long, fixed-interval checks are more likely to land near useful intermediate states, making periodic verification less ill-timed than in shorter-horizon settings. On WebArena, event-driven verification is clearly better: periodic verification is both more expensive and notably less accurate. We believe this is because WebArena episodes are much shorter, with average trajectory length around 10 steps, so fixed checks at every 3, 5, or 7 steps can easily occur too early or too late, missing key milestones or failing to verify the trajectory close enough to the final outcome. In contrast, the event-driven framework adapts to the semantic structure of the trajectory and is therefore better aligned with short, high-variance web tasks.
\begin{table}[t]
\centering
\small
\setlength{\tabcolsep}{5pt}
\begin{tabular}{lccc|ccc}
\toprule
& \multicolumn{3}{c|}{\textbf{OSWorld}} & \multicolumn{3}{c}{\textbf{WebArena}} \\
\cmidrule(lr){2-4} \cmidrule(lr){5-7}
\textbf{Policy} & \textbf{Acc.} & \textbf{Cost/Task} & \textbf{Avg. Step} & \textbf{Acc.} & \textbf{Cost/Task} & \textbf{Avg. Step} \\
\midrule
Periodic-$k$ only   & 55.1\% & \$0.07 & 27.1 & 52.5\% & \$0.24 & 13.3 \\
Stuck + Milestone   & 58.2\% & \$0.05 & 25.2 & 58.8\% & \$0.21 & 12.0 \\
\bottomrule
\end{tabular}
\caption{Comparison of Event-Driven and Periodic Verification Policies. We compare our pipeline against a fixed-interval baseline that invokes the same large model every $k$ steps.}
\label{tab:verification-policy-ablation}
\end{table}

\subsection{Detector Performance}
We evaluate both the milestone and stuck detectors on a held-out 20\% split of the EvoCUA-8B data, with results reported in~\autoref{tab:detector-accuracy}. We also include agreement between independent GPT-5.2 annotation runs on the same set as a reference.
Both learned detectors perform well. The milestone detector achieves 94.1\% accuracy and 62.0\% F1; we emphasize F1 because milestone events are sparse and somewhat ambiguous. These results suggest that compact textual trajectory context is sufficient to recover much of the teacher signal. The stuck detector performs even more strongly, reaching 93.9\% accuracy and 91.5\% F1, consistent with the intuition that stuck behavior is more local and easier to identify from short histories of repeated reasoning and actions.
Taken together, these results support the use of lightweight textual monitors in the event-driven controller. Although they are much cheaper than repeated large-model calls, they provide signals that are accurate enough to support both recovery-oriented escalation and milestone-triggered verification. Representative qualitative stuck and milestone examples are shown in Appendix~\ref{app:detector-examples}.
\begin{table}[h]
\centering
\small
\setlength{\tabcolsep}{6pt}
\begin{tabular}{llcccc}
\toprule
\textbf{Detector} & \textbf{Method} & \textbf{Acc.} & \textbf{Prec.} & \textbf{Rec.} & \textbf{F1} \\
\midrule
Milestone & GPT-5.2 & 94.3\% & 77.1\% & 77.2\% & 77.1\% \\
Milestone & Learned detector               & 94.1\% & 59.4\% & 64.9\% & 62.0\% \\
\midrule
Stuck & GPT-5.2 & 96.5\% & 92.4\% & 93.1\% & 92.7\% \\
Stuck     & Learned detector               & 93.9\% & 91.0\% & 92.0\% & 91.5\% \\
\bottomrule
\end{tabular}
\caption{Detection accuracy of the two monitors on the evaluation split. We report the performance of the learned detectors together with the agreement level of a GPT-5.2 run on the same data.}
\label{tab:detector-accuracy}
\end{table}

\section{Conclusion}

In this paper, we introduced an event-driven, step-level cascade for efficient computer-use agents that allocates expensive large-model inference only when lightweight monitors detect elevated risk. Rather than treating every interaction step as equally difficult, our framework is built on the observation that long-horizon GUI trajectories are highly heterogeneous: many steps are routine and can be handled by a small default policy, while failures tend to concentrate at a limited number of critical moments. By combining a stuck detector for local recovery with a milestone detector for sparse progress verification, the system turns always-on frontier-model inference into adaptive compute allocation over the course of execution. This design is simple, modular, and practical: it can be layered on top of existing computer-use agents without modifying the underlying policy architecture or retraining the stronger model. Across both OSWorld and WebArena, this design consistently improves the performance--efficiency frontier: compared with standalone small models, it recovers large gains in task success, while compared with always-large policies, it retains most of their accuracy at substantially lower latency, cost, and large-model usage. These results suggest that computer-use agent inference should not be treated as uniformly difficult across steps; instead, efficiency comes from identifying the relatively small set of moments where stronger reasoning is most valuable. Overall, our findings position step-level cascading as a practical and modular direction for deploying capable computer-use agents under real-world cost and latency constraints. More broadly, our findings suggest that computer-use agent inference should be viewed as a dynamic systems problem rather than a static per-step prediction problem. Errors often emerge gradually through loops, stalled behavior, or silent drift from the original task intent, and these failure modes call for targeted intervention rather than uniform compute expenditure. Step-level cascading provides one effective way to exploit this structure, showing that fine-grained routing can recover much of the benefit of stronger models without paying their full cost throughout the trajectory.

\newpage

\bibliographystyle{unsrtnat}
\bibliography{main}

@article{dai2025scuba,
  title={SCUBA: Salesforce Computer Use Benchmark},
  author={Dai, Yutong and Ramakrishnan, Krithika and Gu, Jing and Fernandez, Matthew and Luo, Yanqi and Prabhu, Viraj and Hu, Zhenyu and Savarese, Silvio and Xiong, Caiming and Chen, Zeyuan and others},
  journal={arXiv preprint arXiv:2509.26506},
  year={2025}
}

@article{chen2023frugalgpt,
  title={Frugalgpt: How to use large language models while reducing cost and improving performance},
  author={Chen, Lingjiao and Zaharia, Matei and Zou, James},
  journal={arXiv preprint arXiv:2305.05176},
  year={2023}
}

@article{ong2024routellm,
  title={Routellm: Learning to route llms with preference data},
  author={Ong, Isaac and Almahairi, Amjad and Wu, Vincent and Chiang, Wei-Lin and Wu, Tianhao and Gonzalez, Joseph E and Kadous, M Waleed and Stoica, Ion},
  journal={arXiv preprint arXiv:2406.18665},
  year={2024}
}

@article{yue2025masrouter,
  title={Masrouter: Learning to route llms for multi-agent systems},
  author={Yue, Yanwei and Zhang, Guibin and Liu, Boyang and Wan, Guancheng and Wang, Kun and Cheng, Dawei and Qi, Yiyan},
  journal={arXiv preprint arXiv:2502.11133},
  year={2025}
}

@article{abhyankar2025osworld,
  title={OSWorld-Human: Benchmarking the Efficiency of Computer-Use Agents},
  author={Abhyankar, Reyna and Qi, Qi and Zhang, Yiying},
  journal={arXiv preprint arXiv:2506.16042},
  year={2025}
}

@article{awadallah2025fara,
  title={Fara-7B: An Efficient Agentic Model for Computer Use},
  author={Awadallah, Ahmed and Lara, Yash and Magazine, Raghav and Mozannar, Hussein and Nambi, Akshay and Pandya, Yash and Rajeswaran, Aravind and Rosset, Corby and Taymanov, Alexey and Vineet, Vibhav and others},
  journal={arXiv preprint arXiv:2511.19663},
  year={2025}
}

@article{yang2025ferret,
  title={Ferret-ui lite: Lessons from building small on-device gui agents},
  author={Yang, Zhen and Dou, Zi-Yi and Feng, Di and Huang, Forrest and Nguyen, Anh and You, Keen and Attia, Omar and Yang, Yuhao and Feng, Michael and Zhang, Haotian and others},
  journal={arXiv preprint arXiv:2509.26539},
  year={2025}
}

@misc{xie2024osworldbenchmarking,
      title={OSWorld: Benchmarking Multimodal Agents for Open-Ended Tasks in Real Computer Environments}, 
      author={Tianbao Xie and Danyang Zhang and Jixuan Chen and Xiaochuan Li and Siheng Zhao and Ruisheng Cao and Toh Jing Hua and Zhoujun Cheng and Dongchan Shin and Fangyu Lei and Yitao Liu and Yiheng Xu and Shuyan Zhou and Silvio Savarese and Caiming Xiong and Victor Zhong and Tao Yu},
      year={2024},
      eprint={2404.07972},
      archivePrefix={arXiv},
      primaryClass={cs.AI},
      url={https://arxiv.org/abs/2404.07972}, 
}

@article{zhou2023webarena,
  title={Webarena: A realistic web environment for building autonomous agents},
  author={Zhou, Shuyan and Xu, Frank F and Zhu, Hao and Zhou, Xuhui and Lo, Robert and Sridhar, Abishek and Cheng, Xianyi and Ou, Tianyue and Bisk, Yonatan and Fried, Daniel and others},
  journal={arXiv preprint arXiv:2307.13854},
  year={2023}
}

@article{qin2025ui,
  title={Ui-tars: Pioneering automated gui interaction with native agents},
  author={Qin, Yujia and Ye, Yining and Fang, Junjie and Wang, Haoming and Liang, Shihao and Tian, Shizuo and Zhang, Junda and Li, Jiahao and Li, Yunxin and Huang, Shijue and others},
  journal={arXiv preprint arXiv:2501.12326},
  year={2025}
}

@inproceedings{agasheagent,
  title={Agent S: An Open Agentic Framework that Uses Computers Like a Human},
  author={Agashe, Saaket and Han, Jiuzhou and Gan, Shuyu and Yang, Jiachen and Li, Ang and Wang, Xin Eric},
  booktitle={The Thirteenth International Conference on Learning Representations}
}

@article{agashe2025agent,
  title={Agent s2: A compositional generalist-specialist framework for computer use agents},
  author={Agashe, Saaket and Wong, Kyle and Tu, Vincent and Yang, Jiachen and Li, Ang and Wang, Xin Eric},
  journal={arXiv preprint arXiv:2504.00906},
  year={2025}
}

@misc{yang2025qwen3technicalreport,
      title={Qwen3 Technical Report}, 
      author={An Yang and Anfeng Li and Baosong Yang and Beichen Zhang and Binyuan Hui and Bo Zheng and Bowen Yu and Chang Gao and Chengen Huang and Chenxu Lv and Chujie Zheng and Dayiheng Liu and Fan Zhou and Fei Huang and Feng Hu and Hao Ge and Haoran Wei and Huan Lin and Jialong Tang and Jian Yang and Jianhong Tu and Jianwei Zhang and Jianxin Yang and Jiaxi Yang and Jing Zhou and Jingren Zhou and Junyang Lin and Kai Dang and Keqin Bao and Kexin Yang and Le Yu and Lianghao Deng and Mei Li and Mingfeng Xue and Mingze Li and Pei Zhang and Peng Wang and Qin Zhu and Rui Men and Ruize Gao and Shixuan Liu and Shuang Luo and Tianhao Li and Tianyi Tang and Wenbiao Yin and Xingzhang Ren and Xinyu Wang and Xinyu Zhang and Xuancheng Ren and Yang Fan and Yang Su and Yichang Zhang and Yinger Zhang and Yu Wan and Yuqiong Liu and Zekun Wang and Zeyu Cui and Zhenru Zhang and Zhipeng Zhou and Zihan Qiu},
      year={2025},
      eprint={2505.09388},
      archivePrefix={arXiv},
      primaryClass={cs.CL},
      url={https://arxiv.org/abs/2505.09388}, 
}

@misc{xue2026evocuaevolvingcomputeruse,
      title={EvoCUA: Evolving Computer Use Agents via Learning from Scalable Synthetic Experience}, 
      author={Taofeng Xue and Chong Peng and Mianqiu Huang and Linsen Guo and Tiancheng Han and Haozhe Wang and Jianing Wang and Xiaocheng Zhang and Xin Yang and Dengchang Zhao and Jinrui Ding and Xiandi Ma and Yuchen Xie and Peng Pei and Xunliang Cai and Xipeng Qiu},
      year={2026},
      eprint={2601.15876},
      archivePrefix={arXiv},
      primaryClass={cs.AI},
      url={https://arxiv.org/abs/2601.15876}, 
}

@article{xu2024agenttrek,
  title={Agenttrek: Agent trajectory synthesis via guiding replay with web tutorials},
  author={Xu, Yiheng and Lu, Dunjie and Shen, Zhennan and Wang, Junli and Wang, Zekun and Mao, Yuchen and Xiong, Caiming and Yu, Tao},
  journal={arXiv preprint arXiv:2412.09605},
  year={2024}
}

@misc{anthropic_claude_sonnet_45_2025,
  author       = {{Anthropic}},
  title        = {Introducing Claude Sonnet 4.5},
  year         = {2025},
  month        = sep,
  howpublished = {\url{https://www.anthropic.com/news/claude-sonnet-4-5}},
  note         = {Accessed: 2026-03-20}
}

@misc{moonshot_kimi_k25_2026,
  author       = {{Moonshot AI}},
  title        = {Kimi K2.5 -- Kimi API Platform},
  year         = {2026},
  month        = mar,
  howpublished = {\url{https://platform.moonshot.ai/docs/guide/kimi-k2-5-quickstart}},
  note         = {Accessed: 2026-03-20}
}

@misc{openai_gpt_oss_20b_2025,
  author       = {{OpenAI}},
  title        = {gpt-oss-20b Model},
  year         = {2025},
  howpublished = {\url{https://developers.openai.com/api/docs/models/gpt-oss-20b}},
  note         = {Accessed: 2026-03-20}
}

@misc{openai_gpt_5_mini_2025,
  author       = {{OpenAI}},
  title        = {GPT-5 mini Model},
  year         = {2025},
  howpublished = {\url{https://developers.openai.com/api/docs/models/gpt-5-mini}},
  note         = {Accessed: 2026-03-20}
}

@misc{openai_gpt_5_2_2025,
  author       = {{OpenAI}},
  title        = {Introducing GPT-5.2},
  year         = {2025},
  month        = dec,
  howpublished = {\url{https://openai.com/index/introducing-gpt-5-2/}},
  note         = {Accessed: 2026-03-20}
}

@inproceedings{warner2025smarter,
  title={Smarter, better, faster, longer: A modern bidirectional encoder for fast, memory efficient, and long context finetuning and inference},
  author={Warner, Benjamin and Chaffin, Antoine and Clavi{\'e}, Benjamin and Weller, Orion and Hallstr{\"o}m, Oskar and Taghadouini, Said and Gallagher, Alexis and Biswas, Raja and Ladhak, Faisal and Aarsen, Tom and others},
  booktitle={Proceedings of the 63rd Annual Meeting of the Association for Computational Linguistics (Volume 1: Long Papers)},
  pages={2526--2547},
  year={2025}
}

@misc{deng2023mind2webgeneralistagentweb,
      title={Mind2Web: Towards a Generalist Agent for the Web}, 
      author={Xiang Deng and Yu Gu and Boyuan Zheng and Shijie Chen and Samuel Stevens and Boshi Wang and Huan Sun and Yu Su},
      year={2023},
      eprint={2306.06070},
      archivePrefix={arXiv},
      primaryClass={cs.CL},
      url={https://arxiv.org/abs/2306.06070}, 
}

@misc{rawles2025androidworlddynamicbenchmarkingenvironment,
      title={AndroidWorld: A Dynamic Benchmarking Environment for Autonomous Agents}, 
      author={Christopher Rawles and Sarah Clinckemaillie and Yifan Chang and Jonathan Waltz and Gabrielle Lau and Marybeth Fair and Alice Li and William Bishop and Wei Li and Folawiyo Campbell-Ajala and Daniel Toyama and Robert Berry and Divya Tyamagundlu and Timothy Lillicrap and Oriana Riva},
      year={2025},
      eprint={2405.14573},
      archivePrefix={arXiv},
      primaryClass={cs.AI},
      url={https://arxiv.org/abs/2405.14573}, 
}

@inproceedings{nguyen2025gui,
  title={Gui agents: A survey},
  author={Nguyen, Dang and Chen, Jian and Wang, Yu and Wu, Gang and Park, Namyong and Hu, Zhengmian and Lyu, Hanjia and Wu, Junda and Aponte, Ryan and Xia, Yu and others},
  booktitle={Findings of the Association for Computational Linguistics: ACL 2025},
  pages={22522--22538},
  year={2025}
}

@misc{openrouter_pricing,
  author       = {{OpenRouter}},
  title        = {OpenRouter Pricing},
  year         = {2026},
  howpublished = {\url{https://openrouter.ai/pricing}},
  note         = {Accessed: 2026-03-21}
}

@misc{elhattami2025webarena_verified,
  title        = {WebArena Verified},
  author       = {El hattami, Amine and Bredeson, Joel and Thakkar, Megh and LaBouve, Eric and Chapados, Nicolas and Pal, Christopher},
  year         = {2025},
  howpublished = {OpenReview withdrawn submission},
  note         = {Reproducible re-evaluation of the WebArena benchmark},
  url          = {https://openreview.net/forum?id=CSIo4D7xBG}
}

@article{song2025coact,
  title={Coact-1: Computer-using agents with coding as actions},
  author={Song, Linxin and Dai, Yutong and Prabhu, Viraj and Zhang, Jieyu and Shi, Taiwei and Li, Li and Li, Junnan and Savarese, Silvio and Chen, Zeyuan and Zhao, Jieyu and others},
  journal={arXiv preprint arXiv:2508.03923},
  year={2025}
}

@misc{bonatti2024windowsagentarenaevaluating,
      title={Windows Agent Arena: Evaluating Multi-Modal OS Agents at Scale}, 
      author={Rogerio Bonatti and Dan Zhao and Francesco Bonacci and Dillon Dupont and Sara Abdali and Yinheng Li and Yadong Lu and Justin Wagle and Kazuhito Koishida and Arthur Bucker and Lawrence Jang and Zack Hui},
      year={2024},
      eprint={2409.08264},
      archivePrefix={arXiv},
      primaryClass={cs.AI},
      url={https://arxiv.org/abs/2409.08264}, 
}

@article{ding2024hybrid,
  title={Hybrid llm: Cost-efficient and quality-aware query routing},
  author={Ding, Dujian and Mallick, Ankur and Wang, Chi and Sim, Robert and Mukherjee, Subhabrata and Ruhle, Victor and Lakshmanan, Laks VS and Awadallah, Ahmed Hassan},
  journal={arXiv preprint arXiv:2404.14618},
  year={2024}
}

@misc{zhang2026evorouteexperiencedrivenselfroutingllm,
      title={EvoRoute: Experience-Driven Self-Routing LLM Agent Systems}, 
      author={Guibin Zhang and Haiyang Yu and Kaiming Yang and Bingli Wu and Fei Huang and Yongbin Li and Shuicheng Yan},
      year={2026},
      eprint={2601.02695},
      archivePrefix={arXiv},
      primaryClass={cs.CL},
      url={https://arxiv.org/abs/2601.02695}, 
}

@misc{qian2025xroutertrainingcostawarellms,
      title={xRouter: Training Cost-Aware LLMs Orchestration System via Reinforcement Learning}, 
      author={Cheng Qian and Zuxin Liu and Shirley Kokane and Akshara Prabhakar and Jielin Qiu and Haolin Chen and Zhiwei Liu and Heng Ji and Weiran Yao and Shelby Heinecke and Silvio Savarese and Caiming Xiong and Huan Wang},
      year={2025},
      eprint={2510.08439},
      archivePrefix={arXiv},
      primaryClass={cs.LG},
      url={https://arxiv.org/abs/2510.08439}, 
}

@article{wei2026anchor,
  title={ANCHOR: Branch-Point Data Generation for GUI Agents},
  author={Wei, Jinbiao and Zhao, Yilun and Ni, Kangqi and Cohan, Arman},
  journal={arXiv preprint arXiv:2602.07153},
  year={2026}
}

@article{gan2026android,
  title={Android Coach: Improve Online Agentic Training Efficiency with Single State Multiple Actions},
  author={Gan, Guo and Ding, Yuxuan and Chen, Cong and Ren, Yuwei and Huang, Yin and Zhou, Hong},
  journal={arXiv preprint arXiv:2604.07277},
  year={2026}
}

\newpage
\appendix
\section{Details of Benchmarks}
\label{app:benchmarks}

\subsection{OSWorld}

OSWorld~\citep{xie2024osworldbenchmarking} is a real-computer benchmark that wraps full desktop operating systems (primarily Ubuntu) in a controlled virtual machine (VM) environment. Each task is specified by (i) a natural-language instruction describing the user-level goal, (ii) a reproducible VM snapshot encoding the initial desktop state (open applications, files, and window layout), and (iii) an execution script that inspects the final VM state and returns a scalar reward. In our experiments, we follow the benchmark’s execution-based protocol and treat a task as successful if and only if the checker returns a positive signal. The agent interacts with OSWorld exclusively through GUI-level actions: it receives screenshots (and optional structured metadata such as accessibility trees) and issues mouse movements, clicks, scrolls, keyboard events, and window-management operations. We use the standard OSWorld action interface without modifying the underlying environment. Episodes terminate either when the benchmark signals task completion, when the environment reports a failure (e.g., application crash), or when a fixed step budget is reached. We always evaluate on tasks that are disjoint from those used as gold trajectories for data generation.

\subsection{WebArena}
We evaluate on \texttt{WebArena-Verified}, a verified and reproducible release of the original WebArena benchmark \citep{zhou2023webarena, elhattami2025webarena_verified}. WebArena studies long-horizon web-agent execution on realistic browser tasks, while WebArena-Verified preserves the original benchmark setting but strengthens the evaluation pipeline through audited tasks and reference answers, deterministic evaluators, and version-controlled benchmark data. This makes the benchmark substantially more stable and reproducible for comparing agent systems.

In our experiments, we use the ServiceNow \texttt{webarena-verified} release as the evaluation benchmark. Agents interact with the environment through standard browser actions such as clicking, typing, and scrolling, and performance is measured by the benchmark's verified task evaluator. We report both overall success rate and per-domain breakdowns in the main results. Using WebArena-Verified ensures that our reported web-agent results are based on a more reliable and reproducible evaluation protocol than the original benchmark alone.
\section{ModernBERT Fine-tuning Details}
\label{app:training-details}

We fine-tune ModernBERT-base~\citep{warner2025smarter} (\texttt{answerdotai/ModernBERT-base}), a 149M-parameter encoder-only transformer with 22 layers, rotary positional embeddings, and alternating local--global attention, for two binary classification tasks: \textbf{stuck detection} (identifying when agents enter repetitive loops) and \textbf{milestone detection} (identifying steps representing meaningful task progress). A linear classification head is added on top of the \texttt{[CLS]} representation.

Training data is built from 300 GUI-agent trajectories. We randomly generate tasks, execute them with the agent, and collect the resulting trajectories. For each detection, we use GPT-5.2 to label each step. Each trajectory is annotated across five independent runs: a step is labeled positive if it is identified as a milestone or stuck in at least three runs, labeled negative if it is identified in none of the runs, and excluded if it appears in only one or two runs to reduce label noise. For both tasks, each model input consists of the current step together with up to five preceding steps as context. For milestone detection, we additionally prepend the task description.

Both models are trained for 5 epochs with AdamW (learning rate $5 \times 10^{-5}$), per-device batch size 8, maximum sequence length 2{,}048, bf16 mixed precision, and a linear learning rate schedule. Data is split 80/20 for training and evaluation. We also apply inverse-frequency class weighting to the cross-entropy loss. Model selection uses the best F1 score on the test split. Training uses PyTorch 2.4.1 with Hugging Face Transformers and Accelerate.

\section{Detector examples}
\label{app:detector-examples}

\paragraph{Stuck Example 1: Dropdown-closing loop on the Steam page}
\begin{quote}\footnotesize
\textbf{Step 40.} I can see the Dota 2 homepage is displayed, but I need to access the Steam store page to find and add DLCs to the cart. I'll click on the ``Artifact Foundry'' tab to switch back to the Steam store page where I can locate the DLC section.

\textbf{Step 41.} I can see the Artifact Foundry page is currently displayed. I need to scroll down to locate the purchase or add-to-cart button for this DLC, as I haven't found it yet in the visible portion of the page.

\textbf{Step 42.} I can see the page has scrolled down to show customer reviews for Artifact Foundry, but I haven't yet located the purchase or add-to-cart button for the DLC. I need to continue scrolling down to find the section where I can add Artifact Foundry to my cart.

\textbf{Step 43.} I can see a dropdown menu has appeared showing review display options. This isn't helping me find the purchase button for Artifact Foundry. I need to close this dropdown and continue scrolling down the page to locate the add-to-cart button.

\textbf{Step 44.} The dropdown menu for review display options is still open. I need to close this menu by pressing Escape so I can continue scrolling to find the purchase or add-to-cart button for Artifact Foundry.

\textbf{Step 45.} The dropdown menu is still open from the previous action. I need to close this menu by clicking elsewhere on the page so I can continue searching for the purchase or add-to-cart button for Artifact Foundry.
\end{quote}

\textit{Why this is stuck.} The agent spends consecutive steps trying only to dismiss the same dropdown, without returning to the main purchase flow or making progress toward adding the DLC to the cart.

\paragraph{Stuck Example 2: Repetitive cell-selection loop in Calc}
\begin{quote}\footnotesize
\textbf{Step 44.} I've been trying to select cell A4 to continue entering the years for the calculation. I notice I'm still having difficulty with the cell selection. I'll try clicking on cell A4 directly to ensure it's properly selected so I can enter ``2017'' as the next year in the sequence.

\textbf{Step 45.} I notice that I've been repeatedly clicking on cell A4 without successfully entering the year values. I need to pause briefly to allow the system to stabilize and ensure my previous actions have taken effect before proceeding with entering the remaining years.

\textbf{Step 46.} I notice that I've been repeatedly clicking on cell A4 without success. I need to try a different approach to continue entering the years. I'll click on cell A5 to start entering the next year in the sequence.

\textbf{Step 47.} I notice that I've been repeatedly clicking on cell A4 without success. I need to click on cell A5 to continue entering the years for the data series. This will allow me to proceed with filling in the remaining years (2016-2019) for the percentage change calculations.

\textbf{Step 48.} I notice that I've been repeatedly clicking on cell A4 without successfully entering the year values. I need to change my approach by clicking on a different cell to continue entering the years. I'll click on cell A5 to select it for entering the next year.

\textbf{Step 49.} I notice that I've been repeatedly clicking on cell A4 without success. I need to try a different approach to select the correct cell. I'll click on cell A5 to continue entering the years for the data series.
\end{quote}

\textit{Why this is stuck.} The reasoning itself explicitly acknowledges repeated failure, but the behavior still oscillates between the same cell targets (`A4`/`A5`) without entering values or advancing the sheet, which is why we flag it as a stuck loop.

\paragraph{Milestone Example 1: Locating the insertion point in the report}
\begin{quote}\footnotesize
\textbf{Task.} I need to include the experiment results from ``\verb|~/Documents/awesome-desktop/expe-results.xlsx|'' into the currently writing report. Specifically, extract the results of GPT-4 and insert a table into the ``Main Results'' section of my report.

\textbf{Step 2.} I can see the LibreOffice Writer document is open with the report titled ``An Awesome Desktop Environment''. I need to navigate to the ``Main Results'' section to insert the GPT-4 experiment results from the Excel file. I'll scroll down through the document to locate this section.

\textbf{Step 3.} I can see the document structure with sections 1 through 4 visible. I need to continue scrolling down to locate the ``Main Results'' section where I'll insert the GPT-4 experiment results table.

\textbf{Step 4.} I can see the document structure with sections 1 through 4, but I haven't yet located the ``Main Results'' section that needs to receive the GPT-4 experiment results. I need to continue scrolling down to find this section before I can insert the table.

\textbf{Step 5.} I can see the document structure with sections 2, 3, and 4 visible, but I haven't yet located the ``Main Results'' section. I need to continue scrolling down to find where the main results are presented in the report.

\textbf{Step 6.} I can see section 5 ``Benchmarking Advanced Models and Methods'' with subsection 5.1 ``Methods'' visible. I need to continue scrolling down to locate the ``Main Results'' section where I'll insert the GPT-4 experiment results table.

\textbf{Step 7.} I can now see the ``5.2 Main Results'' section in the document. I need to click on this section to position my cursor there so I can insert the table with GPT-4 results from the Excel file.
\end{quote}

\textit{Why this is a milestone.} The agent has successfully reached the correct insertion location in the ``5.2 Main Results'' section, which is a necessary intermediate state before opening the spreadsheet and inserting the GPT-4 results table, so it marks clear task progress even though the edit is not yet finished.

\paragraph{Milestone Example 2: Reaching the final alignment action on slide 3}
\begin{quote}\footnotesize
\textbf{Task.} Align the first textbox on slide 3 to the right, on slide 4 to the center, and on slide 5 to the left. Ensure that the alignment is applied correctly to each respective slide.

\textbf{Step 1.} I can see a LibreOffice Impress presentation with a slide about plants currently displayed. My task is to align textboxes on slides 3, 4, and 5 to the right, center, and left respectively. I need to first navigate to slide 3 to begin the alignment process. I'll start by pausing briefly to ensure the interface is ready for interaction.

\textbf{Step 2.} The interface has stabilized after the initial pause. I can see the slide thumbnails on the left side, including slides 3, 4, and 5 that need to be modified. I'll start by clicking on slide 3 to begin aligning its textbox to the right.

\textbf{Step 3.} I can see slide 3 is now displayed, showing "HELLO THERE!" in red text on the left side. I need to select the first textbox containing "HELLO THERE!" so I can align it to the right as required. I'll click on the text to select it.

\textbf{Step 4.} The textbox on slide 3 is now selected, as indicated by the selection handles around it. I need to access the alignment options to align this textbox to the right. I'll click on the Format menu to find the alignment tools.

\textbf{Step 5.} The Format menu is now open and I can see the "Align Text" option in the menu. I need to click on it to access the text alignment options for the selected textbox on slide 3.

\textbf{Step 6.} The Align Text submenu has appeared with various alignment options. I can see the "Right" option in the submenu, which I need to click to align the text on slide 3 to the right as required by the task.
\end{quote}

\textit{Why this is a milestone.} The agent has successfully reached the exact alignment command needed for the first slide-specific edit, representing clear progress in the multi-slide task without having completed the full assignment.
\section{Prompts}
\label{app:prompts}

\subsection{Prompt Used for Milestone Data Labeling}
\label{app:milestone-labeling-prompt}

\begin{tcolorbox}[promptbox,title=Instruction Prompt]
\begin{PromptText}
You are labeling a GUI agent trajectory to identify milestone steps---steps where meaningful, verifiable progress is achieved.

1. If the trajectory is short/simple, like 5 or 6 steps, you may output only the final step as a milestone.
2. If the trajectory is long, you may list multiple milestones, but don't list too many, and each two milestones should be at least 3 steps apart.
3. If the trajectory becomes stuck (repetition/no progress), ignore steps inside the stuck region unless a milestone occurs later.

Rules:
- A milestone must be meaningful and verifiable from the given step text (action/response/done/fail).
- Prefer higher-level progress markers.
- Do NOT invent UI details you cannot support from the trajectory text.
- For EACH milestone step, provide a clear reasoning explaining why this step represents meaningful progress.
\end{PromptText}
\end{tcolorbox}

\begin{tcolorbox}[promptbox,title=Input Template]
\begin{PromptText}
Task ID: {task_id}
Total steps: {len(trajectory)}

Trajectory:
{trajectory_data}

Return JSON only, matching the schema. For each milestone step, provide the step number AND a reasoning explaining why it's a milestone.
\end{PromptText}
\end{tcolorbox}

\subsection{Prompt Used for Stuck-Step Labeling}
\label{app:stuck-labeling-prompt}

\begin{tcolorbox}[promptbox,title=Instruction Prompt]
\begin{PromptText}
You are an expert AI agent evaluator. Your task is to analyze computer use agent trajectories and identify if the agent got stuck during execution.

An agent is considered "stuck" if:
1. It repeats the same action multiple times without progress
2. It enters an error loop or infinite loop
3. It failed to make meaningful progress for several steps

Analyze the trajectory and return a JSON response in this exact format:
{
  "is_stuck": true/false,
  "stuck_steps": [list of step numbers where agent appears stuck],
  "reasons": [list of reasons explaining why each step is stuck],
  "severity": "low/medium/high",
  "summary": "brief summary of the issue"
}

If the agent is not stuck, return:
{
  "is_stuck": false,
  "stuck_steps": [],
  "reasons": [],
  "severity": "none",
  "summary": "Agent completed task successfully"
}
\end{PromptText}
\end{tcolorbox}

\begin{tcolorbox}[promptbox,title=Input Template]
\begin{PromptText}
Analyze the following agent trajectory for task ID: {task_id}

Total steps: {len(trajectory)}

Trajectory:
{trajectory_data}

Provide your analysis in JSON format as specified.
\end{PromptText}
\end{tcolorbox}

\subsection{Prompt Used for Milestone Verification}
\label{app:milestone-verification-prompt}

\begin{tcolorbox}[promptbox,title=Instruction Prompt]
\begin{PromptText}
You are an expert evaluator of GUI agent progress. Your task is to determine whether the attempted milestone was successfully achieved.

You are given:
1. The task description
2. The actions taken since the previous milestone
3. A BEFORE screenshot from the previous milestone
4. An AFTER screenshot from the current step

Instructions:
- Infer what milestone the agent was attempting to achieve from the task description and the recent actions.
- Compare the BEFORE and AFTER screenshots to determine whether the intended milestone was actually achieved.
- Use the action history as supporting evidence, but base your judgment primarily on whether the AFTER state reflects meaningful progress relative to the BEFORE state.
- Mark "success" as true only if the milestone appears clearly achieved.
- Mark "success" as false if the screenshots do not support that the milestone was completed, if the state is inconsistent with the intended progress, or if the evidence suggests failure.
- Do NOT invent UI details that are not supported by the screenshots or action history.
- For each decision, provide a concise reasoning explaining why the milestone succeeded or failed.
\end{PromptText}
\end{tcolorbox}

\begin{tcolorbox}[promptbox,title=Input Template]
\begin{PromptText}
## Task Description
{task_description}

## Actions Since Previous Milestone
{reasoning_history}

Please analyze whether the milestone step was successful by comparing the BEFORE and AFTER screenshots and examining the actions taken.

Return JSON only, matching this exact schema:
{
  "inferred_milestone": "string - description of what milestone was attempted",
  "success": true/false,
  "reasoning": "string - explanation of why it succeeded or failed"
}
\end{PromptText}
\end{tcolorbox}

\end{document}